# Hybrid Particle Swarm Optimization for Fast and Reliable Parameter Extraction in Thermoreflectance

Bingjia Xiao[1], Tao Chen[1], Wenbin Zhang[2], Xin Qian[1], Puqing Jiang[1,*]

[1]*School of Energy and Power Engineering, Huazhong University of Science and Technology, Wuhan, Hubei 430074, China*

[2]*School of Computing and Information Sciences, Florida International University, Miami, FL, USA*

**ABSTRACT:** Frequency-domain thermoreflectance (FDTR) is a widely used technique for characterizing thermal properties of multilayer thin films. However, extracting multiple parameters from FDTR measurements presents a nonlinear inverse problem due to its high dimensionality and multimodal, non-convex solution space. This study evaluates four popular global optimization algorithms: Genetic Algorithm (GA), Quantum Genetic Algorithm (QGA), Particle Swarm Optimization (PSO), and Fireworks Algorithm (FWA), for extracting parameters from FDTR measurements of a GaN/Si heterostructure. However, none achieve reliable convergence within 60 seconds. To improve convergence speed and accuracy, we propose an AI-driven hybrid optimization framework that combines each global algorithm with a Quasi-Newton local refinement method, resulting in four hybrid variants: HGA, HQGA, HPSO, and HFWA. Among these, HPSO outperforms all other methods, with 80% of trials reaching the target fitness value within 60 seconds, showing greater robustness and a lower risk of premature convergence. In contrast, only 30% of HGA and HQGA trials and 20% of HFWA trials achieve this threshold. We then evaluate the worst-case performance across 100 independent trials for each algorithm when the time is extended to 1000 seconds. Only HPSO, PSO, and HGA consistently reach the target accuracy, with HPSO converging five times faster than the others. HPSO provides a general-purpose solution for inverse problems in thermal metrology and can be readily extended to other model-fitting techniques.

---

# I. Introduction

Thermoreflectance techniques, such as time-domain (TDTR) and frequency-domain thermoreflectance (FDTR), have revolutionized thermal characterization of thin films and bulk materials. [1] [2] These methods fundamentally rely on solving an inverse problem where simulated thermal responses are matched to experimental measurements to extract key parameters such as thermal conductivity, heat capacity, and interfacial thermal conductance. While early implementations depended on manual parameter tuning, the field has progressively adopted optimization algorithms to automate this process. Conventional local optimization methods, including gradient-based approaches like MATLAB's "lsqnonlin" and "fminsearch", offer simplicity for low-dimensional problems but suffer from significant limitations. Their strong dependence on initial guesses and tendency to converge to local minima make them poorly suited for the nonlinear, high-dimensional parameter spaces characteristic of modern thermoreflectance experiments.

The growing complexity of thermoreflectance applications has further exacerbated these challenges. Contemporary studies increasingly require simultaneous extraction of multiple parameters from multifrequency and variable-spot-size measurements, particularly when characterizing novel materials or complex multilayered structures. This introduces intricate coupling between parameters, creating non-unique solutions that challenge conventional optimization approaches. Recent work by Chen and Jiang [3] employed singular value decomposition to reveal that thermoreflectance measurements typically constrain combinations of dimensionless parameters rather than individual properties, explaining why multiple parameter sets can produce equally good fits to experimental data. These findings underscore the critical need for optimization methods capable of robustly navigating multimodal parameter spaces to identify physically meaningful solutions.

Global optimization algorithms have shown promise in addressing these challenges through their ability to explore the entire parameter space without relying on local gradient information. Techniques such as genetic algorithm (GA) [4], quantum genetic algorithm (QGA) [5], particle swarm optimization (PSO) [6] and fireworks algorithm (FWA) [7] have demonstrated success in related thermal applications. For instance, PSO has been effectively applied to thermal management system design [8] and phase-change problems [9], while QGA has shown particular promise in thermoreflectance data analysis [10]. Hybrid approaches combining global exploration with local refinement, such as the QGA-Quasi-Newton method developed by Dong *et al*.[11], have further improved convergence behavior. However, the thermoreflectance community lacks systematic comparisons of these algorithms, leaving open questions about their relative performance for this specific class of problems. Moreover, the computational demands of global optimization methods, particularly QGA when simulated classically, remain a barrier to their widespread adoption in routine measurements.

This study presents a systematic evaluation of global optimization algorithms for thermoreflectance data analysis, with a particular focus on addressing the high-dimensional, nonlinear nature of parameter spaces in multilayer thermal systems. We introduce a hybrid optimization framework, combining the global search capability of PSO with the local refinement of the Quasi-Newton method (HPSO). Our results demonstrate that the HPSO algorithm outperforms standalone global optimization techniques and other hybrid approaches, providing superior accuracy, efficiency, and robustness for reliable thermal property extraction. This framework offers a scalable, practical solution for thermoreflectance and other complex optimization problems in thermal metrology and related fields.

## II. Algorithm Overview

In the realm of optimization, algorithms can be broadly categorized into global and local optimization techniques, each with distinct advantages and limitations. Global optimization algorithms excel in exploring complex, multimodal search spaces by maintaining population diversity and stochastic search mechanisms.

Conversely, local optimization methods focus on refining solutions within a confined neighborhood, leveraging gradient or heuristic-based strategies for rapid convergence.[12] This section systematically compares these two algorithmic paradigms, evaluating their efficacy in terms of solution quality, convergence speed, and robustness across diverse optimization landscapes.

**2.1 Problem formulation**

In thermoreflectance experiments, thermal properties are determined by minimizing the discrepancy between experimental measurements and predictions from a thermal model. This task is formulated as a nonlinear inverse optimization problem, where the goal is to identify the set of unknown thermal parameters, denoted as $\mathbf{X_U} = [k, c_p, G, ...]$, that best fit the experimental data. The objective function is typically defined as the sum of squared errors between the measured signal $S_{\exp}(f_i)$ and the simulated signal $S_{\text{sim}}(\mathbf{X_U}, \mathbf{X_P}, f_i)$ at each modulation frequency $f_i$:

$$\mathcal{F}(\mathbf{X_U}) = \sum_{i=1}^{N} \left( S_{\exp}(f_i) - S_{\text{sim}}(\mathbf{X_U}, \mathbf{X_P}, f_i) \right)^2 \qquad (1)$$

Here, $\mathbf{X_P}$ denotes the set of known input parameters, and $N$ is the total number of data points.

This inverse problem is inherently ill-posed due to several challenges. First, the relationship between thermal properties and the measured signals is highly nonlinear, complicating the optimization process. Second, the simultaneous estimation of multiple parameters increases the dimensionality of the search space, raising the computational burden. Third, the objective function is often non-convex with multiple local minima, which can impede conventional optimization algorithms from reliably identifying the global optimum.

Given these difficulties, the choice of optimization algorithm is critical. Traditional local optimization techniques, such as gradient descent and Newton methods, rely on the differentiability of the objective function and perform well for smooth, convex problems [13]. However, they often fail when applied to highly nonlinear or multimodal functions.

To address these limitations, heuristic global optimization algorithms, such as GA, QGA, PSO, and FWA, have gained considerable attention. These methods do not require gradient information and are well-suited for complex, nonlinear, and ill-posed inverse problems due to their robustness, flexibility, and capability of global search.

The following section provides a comparative overview of these global optimization algorithms and evaluates their performance using benchmark functions before applying them to the thermoreflectance problem.

## 2.2 Global optimization algorithms
### A. Genetic algorithm (GA)

GA is a classical evolutionary optimization method inspired by natural selection. It is effective for solving complex, high-dimensional, and nonlinear optimization problems by evolving a population of candidate solutions through biologically inspired operations: selection, crossover, and mutation [14].

In GA, each candidate solution (chromosome) is encoded as a binary string or a real-valued vector. The algorithm begins with a randomly initialized population and evaluates the fitness of each individual using the objective function. Selection favors fitter individuals, crossover recombines selected chromosomes to produce offspring, and mutation introduces diversity by randomly altering genes to avoid premature convergence.

In this study, thermal parameters are encoded as binary strings and then decoded into real values for fitness evaluation, which is based on the deviation between simulated and observed thermal responses.

GA's population-based and stochastic approach makes it robust in exploring complex error surfaces. However, GA often requires careful tuning of parameters such as population size, crossover rate, and mutation rate. Additionally, GA may converge slowly or become trapped in local minima if population diversity is not maintained.

To address these issues, quantum-inspired variants such as QGA extend GA's capabilities by incorporating quantum computation principles.

**B. Quantum genetic algorithm (QGA)**

QGA enhances classical GA by integrating quantum computing concepts such as qubit representation and quantum rotation gates, thereby improving convergence speed and global search capability [15]. QGA was implemented classically to emulate quantum-inspired search behaviors, given the absence of actual quantum hardware.

Instead of binary or real-valued genes, QGA represents each gene as a qubit, a quantum probabilistic unit expressed as:

$$|\psi\rangle = \alpha|0\rangle + \beta|1\rangle, \text{where } |\alpha|^2 + |\beta|^2 = 1 \quad (2)$$

Each solution (quantum chromosome) consists of a vector of qubits:

$$\mathbf{Q}^t = \begin{bmatrix} \alpha_1^t & \beta_1^t \\ \alpha_2^t & \beta_2^t \\ \vdots & \vdots \\ \alpha_m^t & \beta_m^t \end{bmatrix} \quad (3)$$

where $t$ denotes the iteration index, and $m$ is the number of encoded parameters.

To evaluate fitness, qubits are measured (collapsed) into classical bits according to the probabilities $|\alpha|^2$ and $|\beta|^2$. The resulting binary string is then mapped to a real-valued parameter vector for evaluation using the objective function.

In this study, each thermal parameter is encoded using 20 qubits, each initialized in an equal superposition state ($\alpha = \beta = 1/\sqrt{2}$). This setting ensures uniform sampling of the search space in early generations.

Instead of conventional crossover and mutation, QGA updates qubits using quantum rotation gates:

$$\begin{bmatrix} \alpha' \\ \beta' \end{bmatrix} = \begin{bmatrix} \cos(\Delta\theta) & -\sin(\Delta\theta) \\ \sin(\Delta\theta) & \cos(\Delta\theta) \end{bmatrix} \begin{bmatrix} \alpha \\ \beta \end{bmatrix} \quad (4)$$

Where $\Delta\theta$ is the rotation angle determined based on the fitness of the current solution relative to the global best. An adaptive rotation strategy is employed:

$$\theta' = \theta \left|\frac{\mathcal{F}}{\mathcal{F}_{best}}\right| \tag{5}$$

where $\theta$ is a base rotation angle, $\mathcal{F}$ is the fitness of the current solution, and $\mathcal{F}_{best}$ is the fitness of the best-so-far solution. This approach enables coarse exploration in early stages and fine-grained adjustments as convergence nears.

**C. Particle Swarm Optimization (PSO)**

PSO is a population-based optimization algorithm inspired by the social behavior of birds and fish. Each particle in the swarm represents a candidate solution and navigates the search space by considering its own historical best position and that of the global best [16].

The update rules are:

$$\mathbf{v}_i^{t+1} = \omega \mathbf{v}_i^t + c_1 r_1 (\mathbf{p}_i - \mathbf{x}_i^t) + c_2 r_2 (\mathbf{g} - \mathbf{x}_i^t) \tag{6}$$

$$\mathbf{x}_i^{t+1} = \mathbf{x}_i^t + \mathbf{v}_i^{t+1} \tag{7}$$

Where $\mathbf{x}_i^t$ and $\mathbf{v}_i^t$ are the position and velocity vectors of particle $i$ at iteration $t$. $\mathbf{p}_i$ is the personal best position found by particle $i$, $\mathbf{g}$ is the global best position found by the entire swarm, $\omega$ is the inertia weight controlling the impact of the previous velocity, $c_1$ and $c_2$ are cognitive and social acceleration coefficients, and $r_1$, $r_2$ are random values uniformly distributed in [0, 1]. The balance between exploration and exploitation in PSO is primarily controlled by the parameters $\omega$, $c_1$ and $c_2$. A larger $\omega$ encourages global exploration, while a smaller $\omega$ favors local exploitation.

In this study, each particle encodes a full thermal parameter vector. The fitness of each particle is evaluated via the least-squares error between simulated and experimental data.

PSO's strengths include fast convergence, simplicity, and robustness in multimodal landscapes. It requires minimal parameter tuning and does not depend on gradient information, making it ideal for noisy or discontinuous problems.

**D. Fireworks Algorithm (FWA)**

FWA is inspired by the explosion of fireworks and simulates both local and global search behaviors [7]. Each firework represents a solution and generates sparks in its neighborhood upon explosion.

The number and amplitude of sparks are adaptive: high-quality solutions generate more sparks with smaller amplitudes (local refinement), while lower-quality solutions create fewer, larger-amplitude sparks (exploration). Additionally, Gaussian sparks are occasionally introduced to maintain diversity and avoid stagnation.

In each iteration, all generated sparks (including Gaussian ones) are evaluated, and a new generation of fireworks is selected based on fitness and spatial distribution, maintaining both solution quality and population diversity.

In this study, FWA is applied to the inverse thermal problem, leveraging its dynamic search intensity and ability to balance exploitation and exploration in high-dimensional, nonlinear optimization landscapes.

**E. Performance comparison on a standard benchmark function**

To evaluate and compare the performance of the global optimization algorithms, a benchmark test was conducted using the following standard nonlinear function:

$$Y = X_1 \sin(4\pi X_1) + X_2 \sin(20\pi X_2) \qquad (8)$$

The search domains were defined as $X_1 \in [-3.0, 12.1]$ and $X_2 \in [4.1, 5.8]$. The global minimum of the function, determined via dense grid enumeration, is located at $Y(11.875, 5.775) = -17.65$. A high-resolution surface plot of the function, generated using a 10,000 × 10,000 grid, is shown in Figure 1(a).

Each algorithm (GA, QGA, PSO and FWA) was independently executed 100 times on this benchmark function. The results are summarized in Figure 1(b), where the y-axis shows the ratio of the obtained function value to the global minimum (values closer to 1 indicate higher accuracy), and the x-axis denotes the average optimization time (lower values indicate faster convergence).

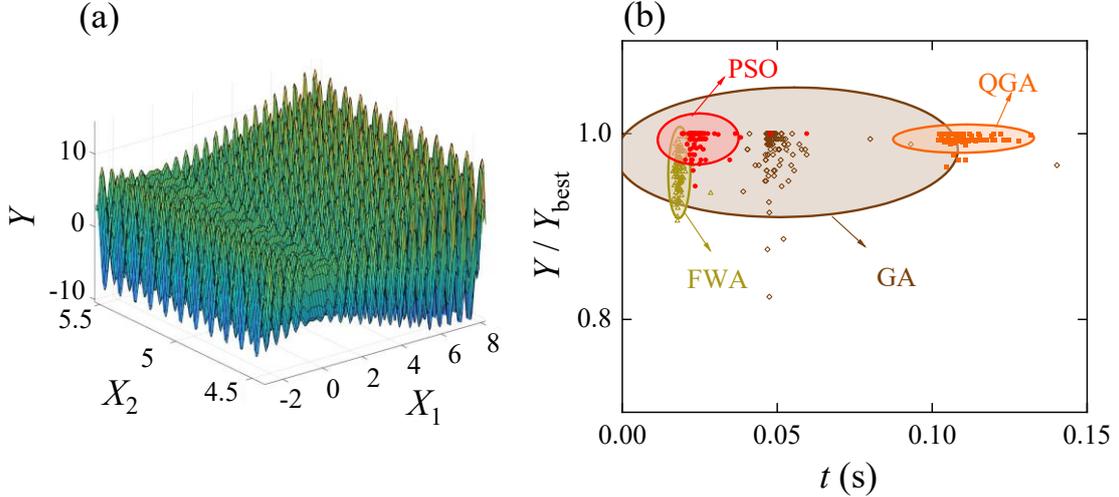

**Fig. 1.** (a) Visualization of the benchmark function $Y = X_1 \sin(4\pi X_1) + X_2 \sin(20\pi X_2)$, plotted over the domain $X_1 \in [-3.0, 12.1]$ and $X_2 \in [4.1, 5.8]$, highlighting the function's complex and multi-modal landscape. (b) Optimization results from 100 independent runs of four global optimization algorithms: Genetic Algorithm (GA), Quantum Genetic Algorithm (QGA), Particle Swarm Optimization (PSO), and Fireworks Algorithm (FWA). The x-axis represents the optimization time, while the y-axis shows the ratio of the obtained function value to the global minimum (with values closer to 1 indicating better performance). Circles indicate the 95% confidence intervals for each algorithm.

The results demonstrate clear differences in algorithm performance. QGA achieved the most accurate solutions, consistently approaching the global minimum, albeit with longer computation times. PSO offered a strong balance between accuracy and speed, showing stable and efficient convergence. FWA optimized quickly but demonstrated greater variability across runs, indicating occasional instability. GA showed moderate results with noticeable variability in convergence time and solution quality.

A detailed comparison of the algorithms' strengths and limitations, based on the repeated optimization trials, is presented in Table 1. Overall, QGA and PSO exhibited the most favorable overall performance in terms of robustness, consistency, and optimization accuracy.

**Table 1**: Comparison of four different global optimization algorithms

| Algorithms | Accuracy | Speed | Stability | Overall Performance |
|---|---|---|---|---|
| QGA | ★★★★★ | ★★☆☆☆ | ★★★★☆ | Excellent |
| PSO | ★★★★☆ | ★★★★☆ | ★★★★☆ | Very Good |
| FWA | ★★★☆☆ | ★★★★★ | ★★☆☆☆ | Moderate |
| GA  | ★★☆☆☆ | ★★☆☆☆ | ★★☆☆☆ | Low |

## 2.3 Local Optimization Methods

While global optimization algorithms such as PSO and QGA are highly effective at locating promising regions in complex, high-dimensional search spaces, they can be computationally intensive and may lack the precision required for fine-tuning solutions. In practice, achieving both broad exploration and precise convergence often requires a combination of global search and efficient local refinement. Therefore, this section evaluates several local optimization techniques that are well suited to this task and discusses their potential use in a hybrid global–local optimization framework.

### A. Quasi-Newton Method

The Quasi-Newton method is a class of optimization algorithms [17] that enhance the classical Newton method by approximating the Hessian matrix $\mathbf{H}$ (or its inverse $\mathbf{H}^{-1}$) to avoid the direct computation of second-order derivatives and matrix inversion. Instead, it constructs an approximate matrix $\mathbf{U}$, or more compactly its inverse $\mathbf{B}^{-1}$, thereby reducing computational complexity and memory usage. Although this approach does not guarantee an optimal search direction at each iteration, the positive definiteness of the approximation ensures overall convergence under standard conditions.

In the classical Newton method, the search direction is computed using the inverse Hessian as

$$\mathbf{d}^{(t)} = -\mathbf{H}_t^{-1}\mathbf{g}_t \tag{9}$$

The Quasi-Newton method avoids computing $\mathbf{H}_t$ directly and instead updates an approximation matrix $\mathbf{U}$ using the Quasi-Newton condition:

$$\Delta x_t = \mathbf{U}_{t+1} \cdot \Delta \mathbf{g}_t \tag{10}$$

Where $\Delta \mathbf{x}_t = \mathbf{x}^{t+1} - \mathbf{x}^t$ and $\Delta \mathbf{g}_t = \mathbf{g}_{t+1} - \mathbf{g}_t$ denotes changes in the position and gradient, respectively.

Several update strategies exist for constructing $\mathbf{U}$, including the Davidon–Fletcher–Powell (DFP) and the Broyden–Fletcher–Goldfarb–Shanno (BFGS) methods. Among these, BFGS is widely used due to its superior convergence rate and numerical stability.

In this study, the Quasi-Newton method is implemented using MATLAB's "fminunc" function, which employs the BFGS algorithm by default. The main steps are as follows:

(1) Initialization: Set the initial point $\mathbf{x}^{(0)}$, convergence tolerance $\varepsilon = 10^{-6}$, initial inverse Hessian approximation $\mathbf{B}_0^{-1}$, and iteration counter $t = 0$.

(2) Compute search direction: calculate the gradient $\mathbf{g}_t = \nabla f(\mathbf{x}^{(t)})$, and set the search direction:

$$\mathbf{d}^{(t)} = -\mathbf{B}_t^{-1} \cdot \mathbf{g}_t \tag{11}$$

(3) Line search: Determine the optimal step size $\lambda_t$ by minimizing the objective function along the search direction:

$$\lambda_t = \underset{\lambda}{\mathrm{argmin}}\, f(\mathbf{x}^{(t)} + \lambda \cdot \mathbf{d}^{(t)}) \tag{12}$$

Then update the current solution:

$$\mathbf{x}^{(t+1)} = \mathbf{x}^{(t)} + \lambda_t \cdot \mathbf{d}^{(t)} \tag{13}$$

(4) Convergence check: If $|\mathbf{g}_{t+1}| < \varepsilon$, the algorithm terminates. Otherwise, continue.

(5) Update inverse Hessian approximation (BFGS formula): Let $\Delta \mathbf{g} = \mathbf{g}_{t+1} - \mathbf{g}_t$, and $\Delta \mathbf{x} = \mathbf{x}^{(t+1)} - \mathbf{x}^{(t)}$. Update $\mathbf{B}^{-1}$ using:

$$\mathbf{B}_{t+1}^{-1} = \left(\mathbf{I} - \frac{\Delta \mathbf{x}_t \Delta \mathbf{g}_t^T}{\Delta \mathbf{x}_t^T \Delta \mathbf{g}_t}\right) \mathbf{B}_t^{-1} \left(\mathbf{I} - \frac{\Delta \mathbf{g}_t \Delta \mathbf{x}_t^T}{\Delta \mathbf{x}_t^T \Delta \mathbf{g}_t}\right) + \frac{\Delta \mathbf{x}_t \Delta \mathbf{x}_t^T}{\Delta \mathbf{x}_t^T \Delta \mathbf{g}_t} \tag{14}$$

where $\mathbf{I}$ is the identity matrix.

(6) Iteration: Increment $t \leftarrow t+1$ and return to Step 2.

**B. Nelder-Mead**

The Nelder-Mead algorithm is a local, derivative-free optimization method for unconstrained problems that does not require gradient information. It is particularly effective for objective functions that are non-differentiable, non-smooth, or expensive to evaluate. The algorithm iteratively updates a simplex, which is a geometric figure consisting of $n+1$ vertices in an *n*-dimensional space. At each iteration, the simplex is modified through one or more of the four operations: reflection, expansion, contraction, and shrinkage, depending on function values at the vertices [18].

While effective in many practical cases, the Nelder–Mead method does not guarantee global optimality, as it lacks exploration mechanisms such as randomness, restart strategies, and population diversity that are characteristic of global optimizers. Additionally, the method is generally unsuitable for high-dimensional problems, where computational cost increases and convergence becomes significantly slower.

In this study, the Nelder-Mead algorithm is implemented using MATLAB's "fminsearch" function, with a function value tolerance of $10^{-6}$ and a maximum of 1000 iterations.

**C. Trust Region Method**

The trust region algorithm (TRA) is a robust numerical method for solving nonlinear optimization problems, particularly in unconstrained settings. Its central idea is to approximate the objective function locally using a simpler surrogate model, typically in the form of a quadratic approximation, within a neighborhood around the current iteration, known as the trust region [19]. Optimization is then carried out on this model, constrained within the trust region.

The size of the trust region is adaptively updated based on the agreement between the model's prediction and the actual objective function. If the model provides an accurate approximation, the trust region is expanded to allow for larger steps; otherwise, it is contracted to improve the reliability of the local model. This adaptive mechanism allows the Trust Region method to break down a complex optimization problem into a series of more manageable local subproblems.

In this study, the Trust Region method is implemented using MATLAB's "lsqnonlin" function from the Optimization Toolbox. The algorithm is configured with a function tolerance of $10^{-6}$, a maximum of 1000 iterations, and the default strategy for dynamically adjusting the trust region radius.

### D. Evaluation of local optimization methods

The Quasi-Newton, Nelder-Mead, and trust region methods each adopt distinct optimization strategies suitable for different types of objective functions. To identify the most appropriate method for integration into a hybrid global-local optimization framework, it is necessary to evaluate their performance under consistent conditions. This section presents a comparative analysis of these three local optimization methods using a standard benchmark function, focusing on convergence behavior, computational efficiency, and solution accuracy.

The benchmark function is defined as:

$$Z(x_1, x_2) = \frac{\sin^2\left(\sqrt{x_1^2 + x_2^2}\right) - 0.5}{(1 + 0.001 \cdot (x_1^2 + x_2^2)^3)^2} - 0.5 \tag{15}$$

Figure 2(a) shows the 3D surface topography of this function, where the peaks and valleys highlight the presence of multiple local extrema. The global minimum is located at (0, 0), with a corresponding function value of $-1$.

To evaluate the convergence performance of the algorithms, two different initial points, (1, 1) and (−2.5, −2.5), were selected. The optimization trajectories from these points are illustrated in Fig. 2(a). All three algorithms converge to the same solution when starting from the same point, demonstrating their sensitivity to initial conditions. When initialized at (1, 1), each algorithm successfully converges to the global minimum located at the center of the surface. However, when initiated at (−2.5, −2.5), all three methods converge to a local minimum located along the circular valley defined by $\sqrt{x_1^2 + x_2^2} \approx 2.83$, where the function value is approximately -0.6772. This behavior highlights the limitation of local optimization methods in multimodal landscapes and supports the need for global search strategies.

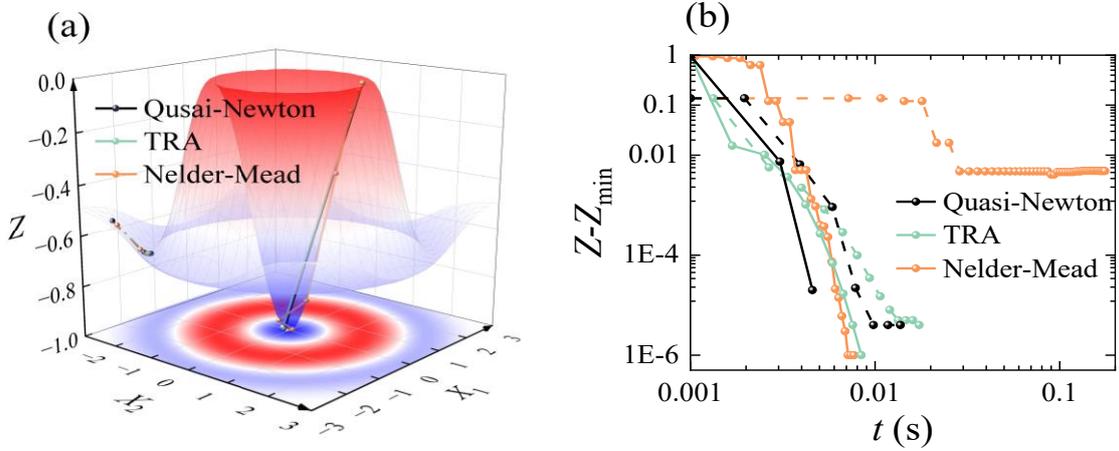

**Fig. 2.** (a) 3D surface plot of the benchmark objective function with superimposed optimization trajectories for the Quasi-Newton (BFGS), Trust Region (TRA), and Nelder-Mead methods. The base plane includes a color-coded heatmap indicating function values, where red represents higher values and blue indicates lower values. Each line traces the path of an algorithm, with markers denoting iteration steps. (b) Comparison of optimization trajectories for the three algorithms, showing the difference between the current function value and the local minimum versus the computational time. The solid trajectory represents the optimization process starting from the initial point (1,1), where the vertical axis denotes the difference between the current objective function value $Z$ and the local optimum $Z_{min} = -1$ (i.e., $Z-Z_{min}$). The dashed trajectory corresponds to the optimization process initialized at (-2.5, -2.5), with the vertical axis representing the difference between the current objective function value $Z$ and another local optimum $Z_{min'} = -0.6772$ (i.e., $Z-Z_{min'}$).

Figure 2(b) compares the computational performance of the three methods under both initial conditions. The lines represent the optimization paths, with symbols marking iteration steps as the algorithms progress toward the local minima. The y-axis shows the difference between the current function value and the respective local minimum (-1 for the initial point (1, 1), and -0.6772 for (-2.5, -2.5)). Results indicate that both the Quasi-Newton method and the trust region approach reliably reach the optimal solutions. The Nelder-Mead method, however, exhibits less consistent accuracy and occasionally fails to converge precisely. In terms of computational time, the trust region method converges rapidly in the early iterations but slows down near the minimum, ultimately requiring more time than the Quasi-Newton method to achieve similar accuracy. Overall, the Quasi-Newton method demonstrates the best balance of convergence speed and accuracy, achieving approximately a 22% reduction in runtime compared to the trust region method.

**2.4 Hybrid optimization strategy**

Global and local optimization methods offer complementary advantages when addressing nonlinear inverse problems, such as extracting thermal properties from thermoreflectance measurements. Local optimizers, including Quasi-Newton and trust region methods, are known for their fast convergence and computational efficiency. However, their performance depends heavily on good initial guesses, and they are susceptible to becoming trapped in local minima. In contrast, global optimization techniques like PSO and QGA are robust to initial conditions and excel at exploring complex, multimodal search spaces. Nonetheless, they can be computationally expensive and less precise in the final stages of convergence.

To leverage the strengths of both approaches, we introduce a hybrid optimization framework that combines global exploration with local refinement. The process begins with a global optimizer that searches broadly across the parameter space to locate promising regions. Once a stopping condition is met, either a fitness threshold or a maximum number of iterations, the algorithm transitions to a local optimizer to fine-tune the solution. This two-stage process enhances convergence speed and accuracy while reducing overall computation time.

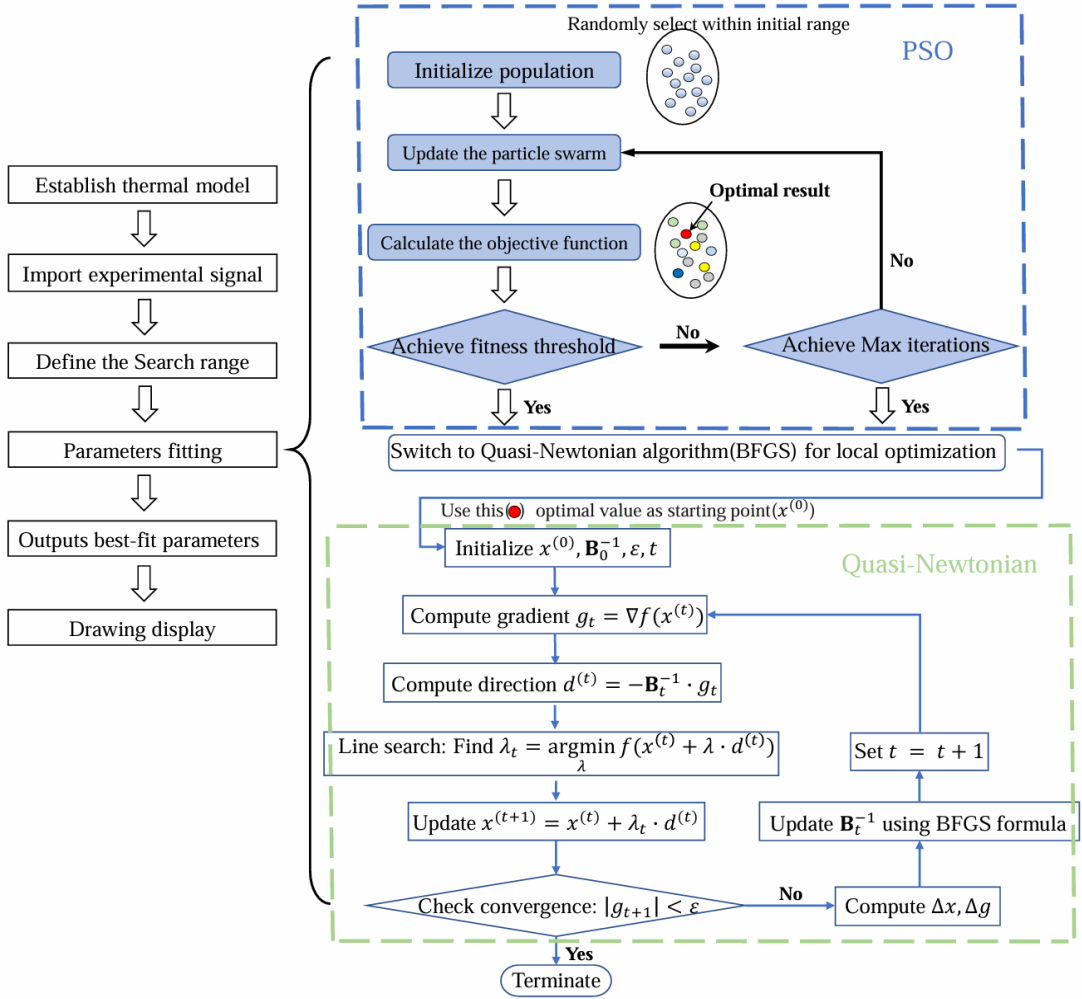

**Fig. 3.** Flow chart of the hybrid PSO-Quasi-Newton (HPSO) optimization algorithm

Figure 3 illustrates the workflow of the hybrid PSO-Quasi-Newton (HPSO) algorithm. In this framework, a three-dimensional heat transfer model simulates the temperature response based on a given parameter set, forming the core of the inverse problem. During the PSO stage, particles are randomly initialized within the search space and iteratively updated according to standard PSO dynamics. Fitness is evaluated by the least-squares deviation between simulated and experimental signals. Once the switching condition is met, the best-performing particle is passed to a Quasi-Newton optimizer, which performs local refinement using the BFGS algorithm to efficiently minimize the objective function.

Similar hybrid strategies are applied to GA, QGA, and FWA, yielding the HGA,

HQGA, and HFWA variants, respectively. By combining the global exploration capabilities of heuristic algorithms with the precision of gradient-based refinement, this hybrid framework effectively addresses the challenges of ill-posed, high-dimensional, and nonlinear inverse problems in thermoreflectance data analysis.

## III. Experimental Setup and Data Collection
### 3.1 Basics of frequency-domain thermoreflectance (FDTR) experiment

As introduced in Section I, FDTR is a non-contact optical technique widely used to measure the thermal properties of materials, especially in micro- and nanoscale heat conduction studies. In an FDTR experiment, as schematically illustrated in Fig. 4, a continuous-wave pump laser, modulated at a known frequency, heats the sample surface periodically. The resulting temperature oscillations cause small changes in the material's reflectance, which are detected by a co-aligned probe laser.

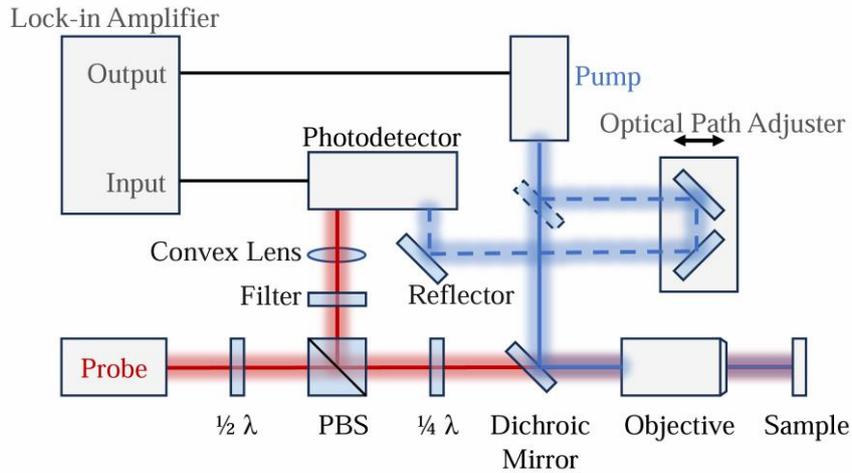

**Fig. 4.** Schematic of the FDTR experimental setup. A modulated continuous-wave pump laser heats the sample, while a co-aligned probe laser detects thermoreflectance signals. A portion of the pump beam is sampled as a reference, with its optical path matched to that of the reflected probe for phase-sensitive detection.

To enable phase-sensitive detection, a portion of the modulated pump beam is sampled and directed to a photodetector as a reference signal. The optical path of this reference is adjusted to match that of the reflected probe beam, allowing accurate measurement of the phase and amplitude of the thermal response.

By sweeping the pump modulation frequency and recording the frequency-dependent phase and amplitude of the reflected probe signal, the surface temperature oscillations are characterized. Fitting the measured response to a multilayer thermal model enables the extraction of key thermophysical parameters of the sample.

**3.2 Data acquisition and parameter identifiability analysis**

Building on the FDTR system described in Section 3.1, we conducted measurements on a GaN/Si heterostructure sample to demonstrate the applicability of the proposed hybrid optimization framework. The sample comprises a 1.08-μm-thick GaN layer grown on a silicon substrate via metal-organic chemical vapor deposition (MOCVD). To mitigate lattice mismatch and thermal stress, intermediate layers consisting of a 290-nm-thick AlN nucleation layer and a 458-nm-thick AlGaN buffer layer were incorporated between the GaN and Si. An 87-nm-thick aluminum (Al) transducer layer was deposited on the GaN surface to enable thermoreflectance-based temperature measurements. The structural composition of the sample is shown in the cross-sectional TEM image in Fig. 5(a).

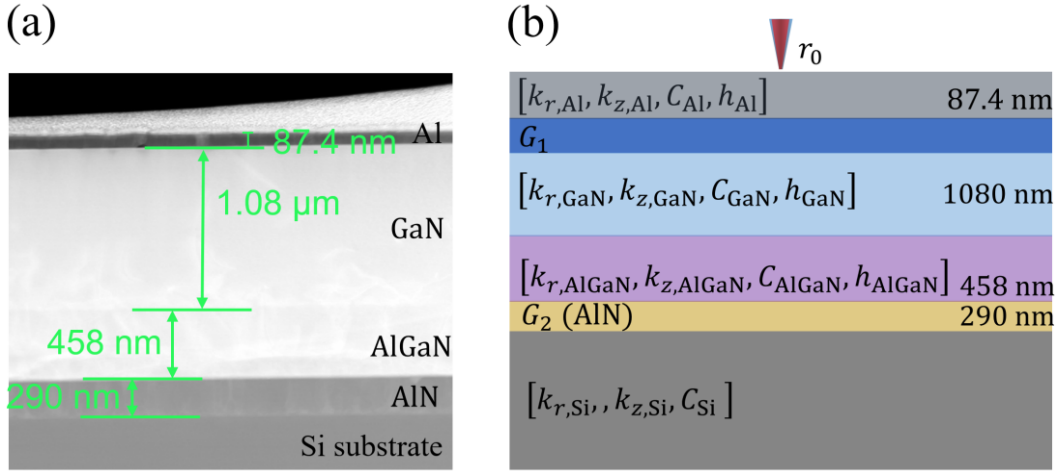

**Fig. 5.** (a) The TEM cross-sectional image shows the structural composition from the silicon substrate upwards, including a 290 nm thick AlN layer, a 458 nm thick AlGaN layer, a 1.08 μm thick GaN layer, and an 87.4 nm thick aluminum layer. (b) The thermal model of the heterostructure displaying the parameters contained in each layer.

A multilayer heat diffusion model was developed to simulate heat transport in the sample under FDTR excitation. The model includes parameters such as in-plane and cross-plane thermal conductivities ($k_r$ and $k_z$), volumetric heat capacity ($C$), thickness ($h$), interfacial thermal conductance ($G$), and laser spot radius ($r_0$). To reduce model complexity and the number of parameters, the following simplifications were applied: (1) Due to weak thermal anisotropy in each layer, all materials were treated as isotropic with $k_r = k_z = k$. (2) The AlN nucleation layer, having a high thermal conductivity and small thickness, was modeled as a thermal interface with interfacial thermal conductance $G_2$, since its thermal resistance is negligible compared to that of the AlGaN buffer, making the FDTR signals not sensitive to the heat capacity of the AlN layer. (3) The thermal resistances of the GaN/AlGaN and AlGaN/AlN interfaces were incorporated into the AlGaN layer, as they are minor relative to the buffer's total resistance. With these simplifications, the number of model parameters was reduced to 14, all labeled in Fig. 5(b).

To assess the identifiability of thermal parameters from the FDTR measurements, we performed a sensitivity analysis based on the logarithmic derivative of the phase signal to each parameter. The sensitivity $S_j$ of the $j$-th parameter $p_j$ is defined as:

$$S_j = \frac{\partial \phi}{\partial \ln p_j} \tag{16}$$

where $\phi$ is the phase signal. This analysis guides the selection of parameters for inversion and helps eliminate those with negligible influence on the signal.

We further applied singular value decomposition (SVD) to assess parameter identifiability and isolate the combinations of parameters most sensitive to the signal. Based on this analysis, five thermal properties were selected for simultaneous extraction: the interfacial thermal conductance $G_1$ at the Al/GaN interface, the heat capacity $C_{\text{GaN}}$ of the GaN layer, the cross-plane thermal conductivities $k_{z,\text{GaN}}$, $k_{z,\text{AlGaN}}$, and $k_{z,\text{Si}}$ of the GaN layer, AlGaN buffer, and Si substrate, respectively.

Table 2 summarizes all parameters used in the thermal model. The five fitting parameters are presented as their respective optimization search ranges, while the remaining input parameters are provided with their nominal values and associated uncertainties. Among the 9 input parameters, the thicknesses $h_{Al}$, $h_{GaN}$, and $h_{AlGaN}$ were measured using TEM, the volumetric heat capacities $C_{Al}$, $C_{AlGaN}$, and $C_{Si}$ were obtained from the literature [3], and the laser spot size $r_0$ was determined using the spatial-domain thermoreflectance (SDTR) method [20]. Moreover, the thermal conductivity of the metal film, $k_{r,Al}$, was calculated from the measured electrical resistivity via the van der Pauw method, using the Wiedemann–Franz law. Since the signals are not sensitive to the interfacial thermal conductance $G_2$ at the AlGaN/AlN/Si interface, it was fixed at $G_2 = 80 \pm 40$ MW/(m² · K), allowing for a wide uncertainty range without affecting the fitting accuracy.

**Table 2.** Summary of thermal model parameters based on the sample structure

| Layer | Parameter | Value/Range | Unit | Source/Method |
|---|---|---|---|---|
| Laser | $r_0$ | 3.4 ± 0.1 and 7.4 ± 0.2 | μm | SDTR |
| Al (transducer) | $k_{Al}$ | 160 ± 16 | W/(m · K) | van der Pauw and Wiedemann-Franz |
| | $C_{Al}$ | 2.44 ± 0.07 | MJ/(m³ · K) | Literature |
| | $h_{Al}$ | 87.4 ± 3 | nm | TEM |
| Al/GaN interface | **$G_1$** | **(10, 300)** | **MW/(m² · K)** | **Fitting** |
| GaN | **$k_{GaN}$** | **(1, 1000)** | **W/(m · K)** | **Fitting** |
| | **$C_{GaN}$** | **(0.5, 5)** | **MJ/(m³ · K)** | **Fitting** |
| | $h_{GaN}$ | 1080 ± 10 | nm | TEM |
| AlGaN buffer | **$k_{AlGaN}$** | **(1, 500)** | **W/(m · K)** | **Fitting** |
| | $C_{AlGaN}$ | 2.6 ± 0.1 | MJ/(m³ · K) | Literature |
| | $h_{AlGaN}$ | 458 ± 8 | nm | TEM |
| AlN (interface layer) | $G_2$ | 80 ± 40 | MW/(m² · K) | Assumed |
| Si substrate | **$k_{Si}$** | **(1, 1000)** | **W/(m · K)** | **Fitting** |
| | $C_{Si}$ | 1.665 ± 0.05 | MJ/(m³ · K) | Literature |

## IV. Results and Discussion

To systematically evaluate the effectiveness of different optimization strategies for extracting thermal parameters from FDTR measurements, we adopt a three-stage analysis approach. First, we assess the reliability of a standalone local optimization method (Quasi-Newton) across 100 independent runs to understand its sensitivity to initial values. Next, we examine the performance of global optimization algorithms (PSO, QGA, GA, FWA) under identical runtime constraints. Finally, we evaluate the performance of hybrid methods that combine global search with local refinement. This progression allows a comprehensive comparison of optimization strategies in terms of robustness, accuracy, and computational efficiency.

### 4.1 Local optimization performance

To evaluate the reliability and sensitivity of local optimization methods, we conducted 100 independent runs of the Quasi-Newton algorithm, using randomly selected initial parameter values within the ranges specified in Table 2. Each run simultaneously fits the experimental FDTR data obtained at two laser spot sizes: 7.4 μm and 3.4 μm. Figure 6(a) presents the distribution of resulting objective function values ($F$), which reveals a wide spread across the 100 trials, indicating significant sensitivity to initial guesses and a high likelihood of convergence to local minima.

To further analyze this variability, we selected three representative cases: the best-performing (Case 1), median (Case 2), and worst-performing (Case 3) results. Figure 6(b) shows the evolution of the objective function over iteration count for each case. When initialized with favorable parameters, the Quasi-Newton method successfully converges to the global optimum (Case 1). However, with less favorable starting points, the algorithm stagnates in local minima (Cases 2 and 3), underscoring its vulnerability to initial conditions.

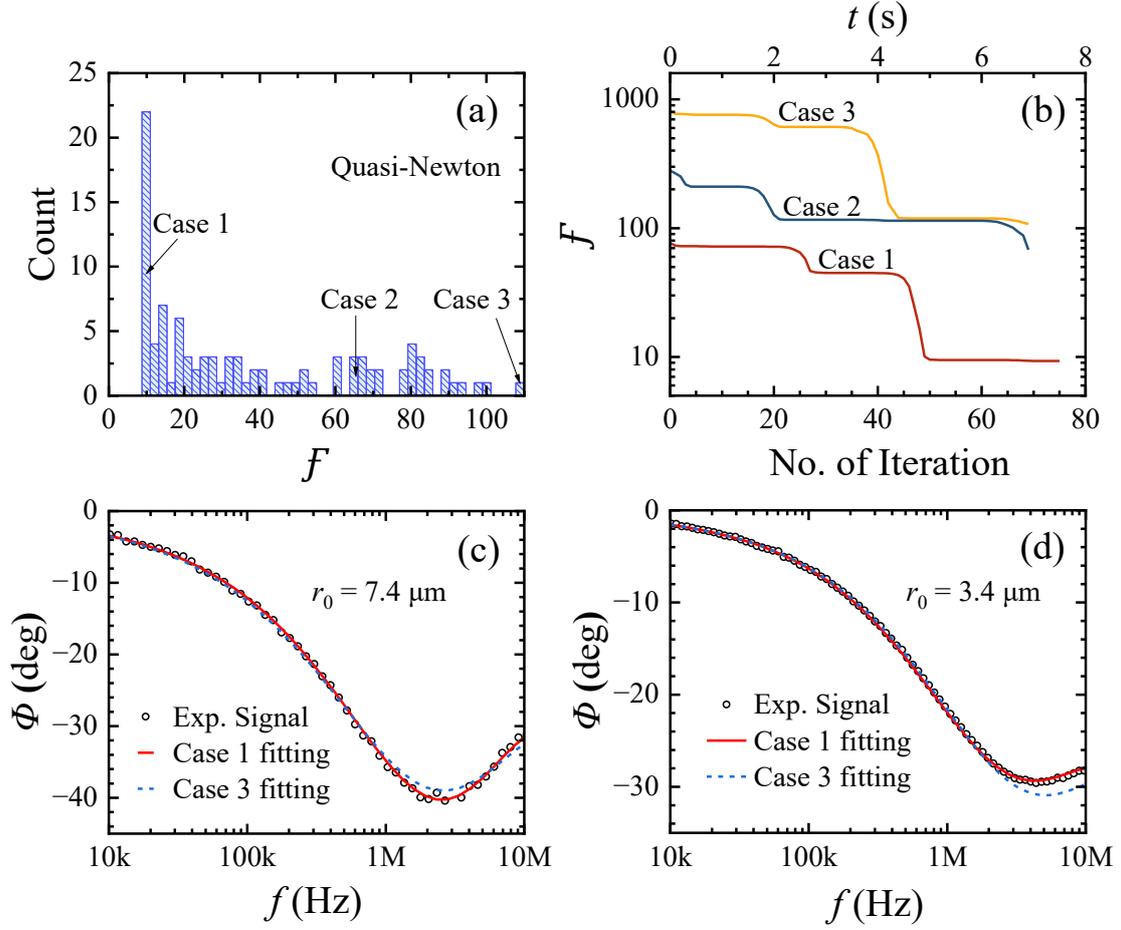

**Fig. 6.** (a) Distribution of objective function values $\mathcal{F}$ from 100 repeated runs of the BFGS algorithm, each initialized with a randomly selected parameter values within the range in Table 2. (b) Convergence trajectories for the best (Case 1), median (Case 2), and worst (Case 3) runs, showing objective function values versus iteration count (bottom x-axis) and optimization time (top x-axis). (c-d) Comparation between FDTR experimental data and simulation results for Case 1 and Case 3 using 7.4 μm spot size and 3.4 μm spot size, respectively.

Figures 6(c) and 6(d) compare the fitting results of Case 1 and Case 3 against the experimental FDTR data acquired using the two laser spot sizes of 7.4 μm and 3.4 μm, respectively, across a frequency range of 10 kHz to 10 MHz. Experimental data are shown as open black circles, while the simulated curves from Case 1 and Case 3 are plotted as solid red and dashed blue lines. As expected, Case 1 yields excellent agreement with the experimental signal, while Case 3 displays noticeable deviations, especially in the high-frequency regions.

These comparisons clearly demonstrate that smaller objective function values correlate with superior fitting performance. Accordingly, all subsequent analyses in this study use the objective function value as the primary quantitative metric for evaluating optimization quality.

**4.2 Global vs. hybrid optimization: convergence, accuracy, and robustness**

To comprehensively evaluate optimization performance, we compared standalone global algorithms (PSO, QGA, GA, FWA) with their corresponding hybrid counterparts that incorporate Quasi-Newton local refinement (HPSO, HQGA, HGA, HFWA). Each algorithm was executed for 100 independent trials. In the hybrid framework, global optimization was allowed to proceed for 60 seconds before switching to the Quasi-Newton method for local refinement. This two-stage strategy aims to balance global exploration and local precision.

Figure 7(a1, b1, c1, d1) summarizes the statistical distributions of optimization outcomes for each method. The red histograms show the objective function values $\mathcal{F}$ obtained after 60 seconds of global optimization alone, while the blue histograms represent the final results after subsequent Quasi-Newton refinement. Among the global methods, PSO achieved the most favorable performance within the 60-second time limit, exhibiting a more concentrated distribution of low objective function values around 100. In contrast, QGA, GA, and FWA have their fitness values distributed between 100 and 1000. After applying the Quasi-Newton refinement, all hybrid variants improved significantly in both convergence and accuracy, with HPSO demonstrating superior robustness—approximately 80% of the runs reached the global optimum. For comparison, there are only 30% of HQGA and HGA trials, and 20% of HFWA trials, reach the same global optimum.

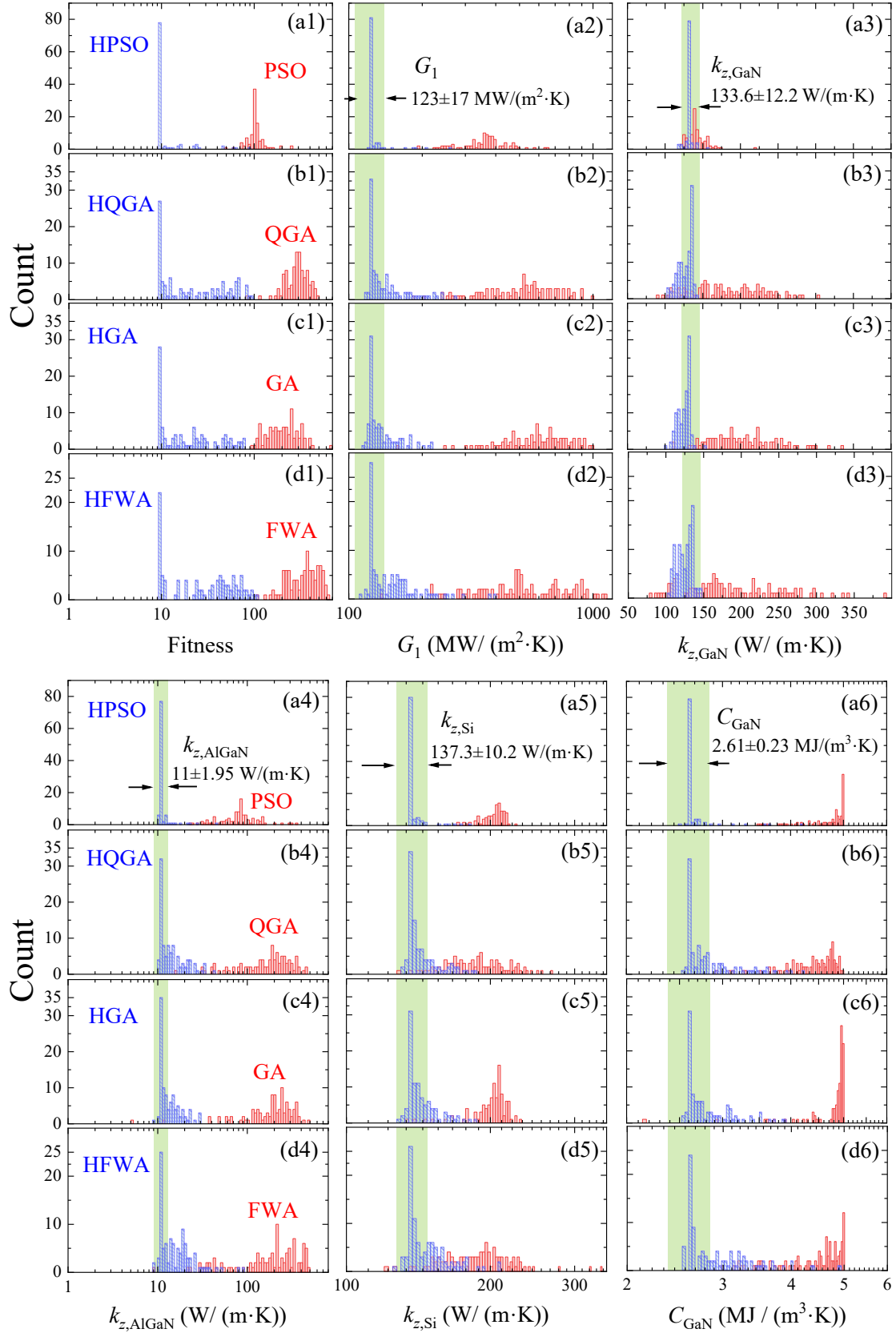

**Fig. 7.** Statistical distributions of parameter estimation results from 100 independent runs of global-local hybrid optimization algorithms. (a1-d1) Objective function values $\mathcal{F}$ after 60 seconds of global optimization (red) and after subsequent Quasi-Newton refinement (blue); (a2-d2) interfacial thermal conductance $G_1$; (a3-d3) cross-plane

thermal conductivity of GaN ($k_{z,\text{GaN}}$); (a4-d4) of AlGaN ($k_{z,\text{AlGaN}}$); (a5-d5) of Si ($k_{z,\text{Si}}$); and (a6-d6) volumetric heat capacity of GaN ($C_{\text{GaN}}$). Green shaded regions indicate theoretical uncertainty ranges of each parameter.

Parameter estimation results further support the effectiveness of hybrid optimization. As shown in subplots (a2)–(a6), none of the pure global algorithms yield the correct outputs. The hybrid strategy significantly improves the accuracy of the outputs. However, only the HPSO-derived parameters closely match literature values and fall entirely within their expected uncertainty bounds, as indicated by the green shaded bands. The other hybrid variants, HGA, HQGA, and HFWA, still have around 50% chance yielding outputs that are outside the uncertainty bands. Considering both optimization efficiency and accuracy, the HPSO method exhibits the best robustness and reliability.

To assess robustness under extended runtime, Figure 8 compares the worst-case convergence trajectories of global-only and hybrid algorithms. Dash-dot curves represent the evolution of objective function values under prolonged global optimization, while solid curves show the results when switching to Quasi-Newton refinement at various time points. Among the standalone global algorithms, PSO again showed the strongest performance, reaching the global optimum within ~300 seconds. In contrast, QGA, GA, and FWA converged much more slowly and plateaued at significantly higher objective function values.

The hybrid algorithms further highlight the benefits of early local refinement. HPSO reached the global optimum within ~60 seconds—nearly five times faster than PSO alone. Among the other hybrid strategies, only HGA eventually converged to the global optimum (after ~300 seconds), while HQGA and HFWA failed to improve beyond suboptimal plateaus. These results emphasize that the effectiveness of hybrid optimization depends not only on the refinement method but also on the global algorithm's ability to provide a sufficiently accurate initial guess before switching.

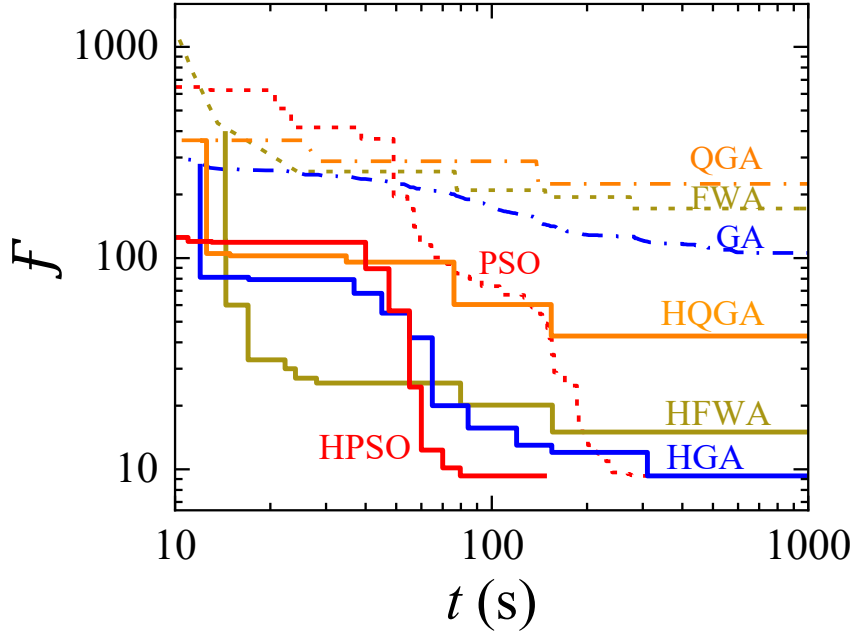

**Fig. 8.** Performance comparison of global-only and global-local hybrid optimization algorithms under varying optimization durations. Dash-dot curves show the objective function value $\mathcal{F}$ achieved by standalone global algorithms over extended runtime. Solid curves represent the corresponding hybrid strategies, where local Quasi-Newton refinement is applied after switching at time $t$.

While QGA have been previously proposed for data processing in transient thermoreflectance (TTR) experiments [5], the poor optimization performance of QGA in this study can be attributed to its computational overhead when implemented on classical computers. Simulating quantum bit superposition and evolution processes demands frequent quantum state measurement, decoding, and fitness evaluation, significantly increasing convergence time. This makes QGA less efficient than PSO, as it fails to reach optimal solutions within the same computational budget. Furthermore, the suboptimal initial solutions generated by QGA hinder the performance of subsequent local optimization (HQGA), leading to poorer fitting accuracy compared to the HPSO approach, as shown by our experimental results.

Thus, hybrid optimization algorithms, such as HPSO, provide a more efficient and accurate solution to thermoreflectance inverse problems by combining the exploratory power of PSO with the local refinement of Quasi-Newton methods, ensuring both speed and precision.

## V. Conclusion

This study presents a hybrid optimization framework that addresses the nonlinear, high-dimensional inverse problem in thermoreflectance-based thermal property extraction. By combining the global search capability of Particle Swarm Optimization (PSO) with the local refinement of the Quasi-Newton method, the proposed HPSO algorithm achieves significantly improved convergence speed, accuracy, and robustness compared to standalone global or local methods. In 100 independent trials on FDTR data from a multilayer GaN/Si heterostructure, HPSO achieved the highest success rate in reaching the target fitness threshold within 60 seconds and was the only method to consistently yield thermal parameter estimates within established uncertainty ranges. These results demonstrate the strong potential of hybrid algorithms like HPSO for reliable and efficient parameter estimation in thermal metrology. The framework is broadly applicable to other inverse problems, including techniques such as TDTR and 3ω, with potential for use in energy systems, electronic packaging, and thermal material characterization.

**Acknowledgment**

X.Q. acknowledges the funding support from National Key R&D program (2022YFA1203100). P.J. acknowledges support from the National Natural Science Foundation of China (NSFC) through Grant No. 52376058.